\documentclass[a4paper, times, 10pt,twocolumn]{article}
\usepackage[top=4.9cm,bottom=3.7cm,left=1.5cm,right=1.5cm]{geometry}
\usepackage{ICMLC}
\usepackage{times}
\usepackage{graphicx}
\usepackage{indentfirst}
\usepackage{latexsym}
\usepackage{amsfonts}
\usepackage{amssymb,amsmath}
\usepackage{bm}

\begin{document}

\title{INFINITE MIXTURES OF MULTIVARIATE GAUSSIAN PROCESSES}  

\author{\bf{\normalsize{SHILIANG SUN}}\\ 
\\
\normalsize{Department of Computer Science and Technology, East China Normal University}\\
\normalsize{500 Dongchuan Road, Shanghai 200241, China}\\
\normalsize{E-MAIL: slsun@cs.ecnu.edu.cn, shiliangsun@gmail.com}\\
\\}

\maketitle \thispagestyle{empty}

\begin{abstract}
   {This paper presents a new model called infinite mixtures of multivariate Gaussian processes, which can be
   used to learn vector-valued functions and applied to multitask learning. As an extension of the single multivariate
   Gaussian process, the mixture model has the advantages of modeling multimodal data and alleviating the
   computationally cubic complexity of the multivariate Gaussian
   process. A Dirichlet process prior is adopted to allow the (possibly infinite)
   number of mixture components to be automatically inferred from
   training data, and Markov chain Monte Carlo sampling techniques are used for
   parameter and latent variable inference. Preliminary experimental
   results on multivariate regression show the feasibility of the proposed model.}
\end{abstract}

\setlength{\arraycolsep}{0.1em}
\Section{Introduction}

Gaussian processes provide a principled probabilistic approach to
pattern recognition and machine learning. Formally, a Gaussian
process is a collection of random variables such that any finite
number of them obey a joint Gaussian prior distribution. As a
Bayesian nonparametric model, the Gaussian process model proves to
be very powerful for general function learning problems such as
regression and classification~\cite{Rasmussen06gpml,Sun11viIMGP}.

Recently, motivated by the need to learn vector-valued functions and
for multitask learning, research on multivariate or multi-output
Gaussian processes has attracted a lot of attention. By learning
multiple related tasks jointly, the common knowledge underlying
different tasks can be shared, and thus a performance gain is likely
to be obtained~\cite{Ji13M2SVM}. Representative works on
multivariate Gaussian processes include the methods given
in~\cite{Bonilla08MtGP,Yuan11CMoR,Alvarez11CMoGP}.

However, it is well known that Gaussian processes suffer from two
important limitations~\cite{Sun11viIMGP,Rasmussen02Infm}. First,
limited by the inherent unimodality of Gaussian distributions,
Gaussian processes cannot characterize multimodal data which are
prevalent in practice. Second, they are computationally infeasible
for big data, since inference requires the inversion of an $N\times
N$ and $NM\times NM$ covariance matrix respectively for a
single-variate and multivariate Gaussian process, where $N$ is the
number of training examples and $M$ is the output dimensionality.

These two limitations can be greatly alleviated by making use of
mixtures of Gaussian processes~\cite{Tresp01MGP} where there are
multiple Gaussian processes to jointly explain data and one example
only belongs to one Gaussian process component. For mixtures of
Gaussian processes, the infinite mixtures based on Dirichlet
processes~\cite{Teh10DP} are prevailing because they permit the
number of components to be inferred directly from data and thus
bypass the difficult model selection problem on the component
number.

For single-variate or single-output Gaussian processes, there were
already some variants and implementations for infinite mixtures
which brought great success for data modeling and prediction
applications~\cite{Sun11viIMGP,Rasmussen02Infm,Meeds06IM}. However,
no extension of multivariate Gaussian processes to mixture models
has been presented yet. Here, we will fill this gap by proposing an
infinite mixture model of multivariate Gaussian processes. It should
be noted that the implementation of this infinite model is very
challenging because the multivariate Gaussian processes are much
more complicated than the single-variate Gaussian processes.

The rest of this paper is organized as follows. After providing the
new infinite mixture model in Section~\ref{sec:Model}, we show how
the hidden variable inference and prediction problems are performed
in Section~\ref{sec:Inference} and Section~\ref{sec:Prediction},
respectively. Then, we report experimental results on multivariate
regression in Section~\ref{sec:Expe}. Finally, concluding remarks
and future work directions are given in Section~\ref{sec:Conc}.

\Section{The proposed model}
\label{sec:Model}

\begin{figure}[th]
  \centering
  \includegraphics[width=235pt]{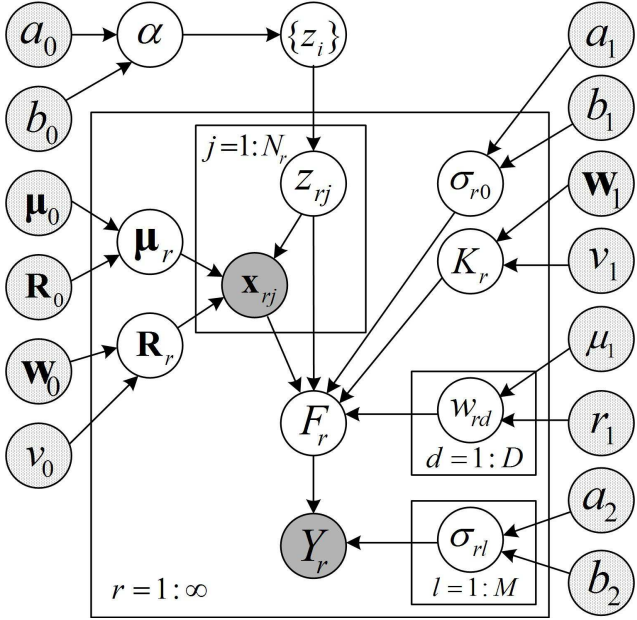}
  \caption{The graphical model for IMMGP}
  \label{GM_IMMGP}
\end{figure}

The graphical model for the proposed infinite mixture of
multivariate Gaussian processes (IMMGP) on the observed training
data $\mathcal{D}=\{\textbf{x}_i, \textbf{y}_i\}_{i=1}^N$ is
depicted in Figure~\ref{GM_IMMGP}.

In the graphical model, $r$ indexes the $r$th Gaussian process
component in the mixture, which can be infinitely large if enough
data are provided. $N_r$ is the number of examples belonging to the
$r$th component. $D$ and $M$ are the dimensions for the input and
output space, respectively. The set $\{\alpha, \{\bm\mu_r\},
\{\textbf{R}_r\}, \{\sigma_{r0}\}, \{K_r\}, \{w_{rd}\},
\{\sigma_{r\ell}\}\}$ includes all random parameters, which is
denoted here by $\Theta$. The latent variables are $z_i$
$(i=1,\ldots,N)$ and $F_r$ $(r=1:\infty)$, where $F_r$ can be
removed from the graphical model by integration if we directly
consider a distribution over $\{Y_r\}$. Denote the set of latent
indicators by $Z$, that is, $Z=\{z_i\}_{i=1}^N$. Since $F_r$ is for
illustrative purposes only, the latent indicators $Z$ and random
parameters $\Theta$ constitute the total hidden variables. The
circles in the left and right columns of the graphical model
indicate the hyperparameters, whose values are usually found by
maximum likelihood estimation or designated manually if people have
a strong belief on them.

The observation likelihood for our IMMGP is
\begin{eqnarray}
&& p(\{\textbf{x}_i, \textbf{y}_i\}|\Theta) \nonumber\\
&=&\sum_Z p(Z|\Theta) \prod_r p(\{\textbf{y}_i:
z_i=r\}|\{\textbf{x}_i: z_i=r\}, \Theta) \times \nonumber\\
&& p(\{\textbf{x}_i: z_i=r\}|
\Theta) \nonumber\\
&=&\sum_Z p(Z|\Theta) \prod_r p(\{\textbf{y}_i:
z_i=r\}|\{\textbf{x}_i: z_i=r\}, \Theta) \times \nonumber\\
&&\prod_{j=1}^{N_r} p(\textbf{x}_{rj}|\bm\mu_r, \textbf{R}_r) .
\end{eqnarray}

\SubSection{Distributions for hidden variables}

$\alpha$ is the concentration parameter of the Dirichlet process,
which controls the prior probability of assigning an example to a
new mixture component and thus influences the total number of
components in the mixture model. A gamma distribution
$\mathcal{G}(\alpha|a_0, b_0)$ is used. We use the parameterization
for the gamma distribution given in~\cite{Bishop06prml}. Given
$\alpha$ and $\{z_i\}_{i=1}^n$, the distribution of $z_{n+1}$ is
easy to get with the Chinese restaurant process
metaphor~\cite{Teh10DP}.

The distribution over the input space for a mixture component is
given by a Gaussian distribution with a full covariance
\begin{equation}
p(\textbf{x}| z=r, \bm\mu_r, \mathbf{R}_r)=\mathcal
{N}(\textbf{x}|\bm\mu_r, \mathbf{R}_r^{-1}) ,
\end{equation}
where $\mathbf{R}_r$ is the precision (inverse covariance) matrix.
This input model is often flexible enough to provide a good
performance, though people can consider to adopt mixtures of
Gaussian distributions to model the input space. Parameters
$\bm\mu_r$ and $\mathbf{R}_r$ are further specified by a Gaussian
distribution prior and a Wishart distribution prior, respectively
\begin{equation}
\bm\mu_r \sim \mathcal {N}(\bm\mu_0, \mathbf{R}_0^{-1}), \;
\mathbf{R}_r \sim  \mathcal {W}(\mathbf{W}_0, \nu_0) .
\end{equation}
The parameterization for the Wishart distribution is the same as
that in~\cite{Bishop06prml}.

A Gaussian process prior is placed over the latent functions
$\{f_{r\ell}\}_{\ell=1}^M$ for component $r$ in our model. Assuming
the Gaussian processes have zero mean values, we set
\begin{eqnarray}
\label{eqnGPprior} \mathbb{E}\big(f_{r\ell}(\textbf{x}) f_{r
k}(\textbf{x}^\prime)\big) &=&\sigma_{r0} K_r(\ell, k)
k_r(\textbf{x},\textbf{x}^\prime), \nonumber\\
y_{r\ell}(\textbf{x}) &\sim& \mathcal{N} (f_{r\ell}(\textbf{x}),
\sigma_{r\ell}),
\end{eqnarray}
where scaling parameter $\sigma_{r0}>0$,  $K_r$ is a positive
semi-definite matrix that specifies the inter-task similarities,
$k_r(\cdot,\cdot)$ is a covariance function over inputs, and
$\sigma_{r\ell}$ is the noise variance for the $\ell$th output of
the $r$th component. The prior of the $M \times M$ positive
semi-definite matrix $K_r$ is given by a Wishart distribution
$\mathcal {W}(\mathbf{W}_1, \nu_1)$. $\sigma_{r0}$ and
$\sigma_{r\ell}$ are given gamma priors
$\mathcal{G}(\sigma_{r0}|a_1, b_1)$ and
$\mathcal{G}(\sigma_{r\ell}|a_2, b_2)$, respectively. We set
\begin{equation}
k_r(\textbf{x},\textbf{x}^\prime)=\exp\Big(-\frac{1}{2}\sum\nolimits_{d=1}^D
w^2_{rd}(\textbf{x}_d - \textbf{x}^\prime_d)^2\Big) ,
\end{equation}
where $w_{rd}$ obeys a log-normal distribution $\mathcal{N}(\ln
w_{rd}|\mu_1, r_1)$ with mean $\mu_1$ and variance $r_1$. The whole
setup for a single Gaussian process component is in large difference
with that in~\cite{Bonilla08MtGP}.

\Section{Inference}
\label{sec:Inference}

Since exact inference on the distribution $p(Z, \Theta|\mathcal{D})$
is infeasible, in this paper we use Markov chain Monte Carlo
sampling techniques to obtain $L$ samples $\{Z_j,
\Theta_j\}_{j=1}^L$ to approximate the distribution $p(Z,
\Theta|\mathcal{D})$.

In particular, Gibbs sampling is adopted to represent the posterior
of the hidden variables. First of all, we initialize all the
variables in $\{Z, \Theta\}$ by sampling them from their priors.
Then the variables are updated using the following steps.
\begin{enumerate}
\item[(1)] Update indicator variables $\{z_i\}_{i=1}^N$ one by one, by cycling through the training data.
\item[(2)] Update input and output space Gaussian process parameters
 $\{\{\bm\mu_r\}, \{\textbf{R}_r\}, \{\sigma_{r0}\}, \{K_r\},
\{w_{rd}\}, \{\sigma_{r\ell}\}\}$ for each Gaussian process
component in turn.
\item[(3)] Update Dirichlet process concentration parameter $\alpha$.
\end{enumerate}

These three steps constitute a Gibbs sampling sweep over all hidden
variables, which are repeated until the Markov chain has adequate
samples. Note that samples in the burn-in stage should be removed
from the Markov chain and are not used for approximating the
posterior distribution.

In the following subsections, we  provide the specific sampling
method and formulations involved for each update.

\SubSection{Updating indicator variables}

Let $Z_{-i}=Z \backslash z_i=\{z_1, \ldots, z_{i-1}, z_{i+1},
\ldots, z_N\}$ and $\mathcal{D}_{-i}=\mathcal{D} \backslash
\{\textbf{x}_i, \textbf{y}_i\}$. To sample $z_i$, we need the
following posterior conditional distribution
\begin{eqnarray}
\label{eqnUpdIV}
&& p(z_i|Z_{-i}, \Theta, \mathcal{D}) \nonumber\\
&\propto& p(z_i|Z_{-i}, \Theta)
p(\mathcal{D}|z_i, Z_{-i}, \Theta) \nonumber\\
&\propto& p(z_i|Z_{-i}, \Theta) p\left(\textbf{y}_i|\{\textbf{y}_j:
j\neq i, z_j=z_i\}, \{\textbf{x}_j: z_j=z_i\},
\Theta\right) \nonumber\\
&& \times p(\textbf{x}_i|\bm\mu_{z_i}, \textbf{R}_{z_i}) ,
\end{eqnarray}
where we have used a clear decomposition between the joint
distributions of $\{\textbf{x}_i, \textbf{y}_i\}$ and
$\mathcal{D}_{-i}$.

It is not difficult to calculate the three terms involved in the
last two lines of (\ref{eqnUpdIV}). However, the computation of
$p(\textbf{y}_i|\{\textbf{y}_j: j\neq i, z_j=z_i\}, \{\textbf{x}_j:
z_j=z_i\}, \Theta)$ may be more efficient if some approximation
scheme or acceleration method is adopted. In addition, for exploring
new experts, we just sample the parameters once from the prior, use
them for the new expert, and then calculate (\ref{eqnUpdIV}),
following \cite{Meeds06IM,Rasmussen00IGMM}. The indicator variable
update method is also algorithm 8 from \cite{Neal98Sampling} with
the auxiliary component parameter $m=1$.

\SubSection{Updating input space component parameters}

For the input space parameters $\bm\mu_r$ and $\textbf{R}_r$, they
can be sampled directly because their posterior conditional
distributions have a simple formulation as a result of using
conjugate priors.
\begin{eqnarray}
&&p(\bm\mu_r| Z, \Theta \backslash \bm\mu_r, \mathcal{D})
= p(\bm\mu_r|\{\textbf{x}_{rj}\}_{j=1}^{N_r}, \textbf{R}_r)\nonumber\\
&\propto& p(\bm\mu_r) p(\{\textbf{x}_{rj}\}_{j=1}^{N_r}|  \bm\mu_r,
\textbf{R}_r)\nonumber\\
&\propto& |\textbf{R}_0|^{1/2} \exp
\{-\frac{1}{2}(\bm\mu_r-\bm\mu_0)^\top\textbf{R}_0
(\bm\mu_r-\bm\mu_0)\} \times\nonumber\\
&& \prod_j |\textbf{R}_r|^{1/2} \exp\{-\frac{1}{2}
(\textbf{x}_{rj}-\bm\mu_r)^\top\textbf{R}_r
(\textbf{x}_{rj}-\bm\mu_r)\}\nonumber\\
&\propto& \exp\{-\frac{1}{2}[\bm\mu_r^\top \textbf{R}_0 \bm\mu_r- 2
\bm\mu_r^\top \textbf{R}_0 \bm\mu_0 + \nonumber\\
&&\quad \sum\nolimits_j (\bm\mu_r^\top \textbf{R}_r \bm\mu_r- 2
\bm\mu_r^\top \textbf{R}_r \textbf{x}_{rj})]\} ,
\end{eqnarray}
and therefore
\begin{eqnarray}
&&p(\bm\mu_r| Z, \Theta \backslash \bm\mu_r, \mathcal{D})
\nonumber\\
&=& \mathcal{N}((\textbf{R}_0 + N_r \textbf{R}_r)^{-1}
(\textbf{R}_0\bm\mu_0 + \textbf{R}_r \sum\nolimits_j \textbf{x}_j),
(\textbf{R}_0 + N_r \textbf{R}_r)^{-1}) . \nonumber
\end{eqnarray}

\begin{eqnarray}
&&p(\textbf{R}_r| Z, \Theta \backslash \textbf{R}_r, \mathcal{D})
= p(\textbf{R}_r|\{\textbf{x}_{rj}\}_{j=1}^{N_r}, \bm\mu_r)\nonumber\\
&\propto& p(\textbf{R}_r) p(\{\textbf{x}_{rj}\}_{j=1}^{N_r}|
\bm\mu_r,
\textbf{R}_r)\nonumber\\
&\propto& |\textbf{R}_r|^{(\nu_0 - D - 1)/2} \exp
\{-\frac{1}{2}\mbox{Tr}(\textbf{W}_0^{-1} \textbf{R}_r)\}
\times \nonumber\\
&&\prod_j |\textbf{R}_r|^{1/2} \exp\{-\frac{1}{2}
(\textbf{x}_{rj}-\bm\mu_r)^\top\textbf{R}_r
(\textbf{x}_{rj}-\bm\mu_r)\} \nonumber\\
&\propto& |\textbf{R}_r|^{(\nu_0 + N_r - D - 1)/2}
\times\nonumber\\
&&\exp \{-\frac{1}{2}\mbox{Tr}((\textbf{W}_0^{-1} + \sum\nolimits_j
(\textbf{x}_{rj}-\bm\mu_r)(\textbf{x}_{rj}-\bm\mu_r)^\top)\textbf{R}_r)\},
\nonumber
\end{eqnarray}
and thus
\begin{eqnarray}
&&p(\textbf{R}_r| Z, \Theta \backslash \textbf{R}_r,
\mathcal{D})\nonumber\\
&=& \mathcal{W}\left(\big(\textbf{W}_0^{-1} + \sum\nolimits_j
(\textbf{x}_{rj}-\bm\mu_r)(\textbf{x}_{rj}-\bm\mu_r)^\top\big)^{-1},
\; \nu_0 + N_r\right) . \nonumber
\end{eqnarray}

\SubSection{Updating output space component parameters}

Note that $Y_r =\{\textbf{y}_i: 1\leq i \leq N, z_i=r\}$ and
$|Y_r|=N_r$. In this subsection, we denote its $N_r$ elements by
$\{\textbf{y}_r^j\}_{j=1}^{N_r}$ which correspond to
$\{\textbf{x}_{r}^{j}\}_{j=1}^{N_r}$.

Define the complete $M$ outputs in the $r$th GP as
\begin{equation}
\textbf{y}_r =(y_{r1}^1, \ldots, y_{r1}^{N_r}, y_{r2}^1, \ldots,
y_{r2}^{N_r}, \ldots, y_{rM}^1, \ldots, y_{rM}^{N_r})^\top ,
\end{equation}
where $y_{r\ell}^j$ is the observation for the $\ell$th output on
the $j$th input. According to the Gaussian process assumption given
in (\ref{eqnGPprior}), the observation $\textbf{y}_r$ follows a
Gaussian distribution
\begin{equation}
\label{eqnLikeOutput} \textbf{y}_r \sim \mathcal{N}(\textbf{0},
\Sigma), \quad \Sigma=\sigma_{r0} K_r \otimes K_r^x + D_r \otimes I
,
\end{equation}
where $\otimes$ denotes the Kronecker product, $K_r^x$ is the $N_r
\times N_r$ covariance matrix between inputs with
$K_r^x(i,j)=k_r(\textbf{x}_{r}^i,\textbf{x}_r^j)$, $D_r$ is an
$M\times M$ diagonal matrix with $D_r(i,i)=\sigma_{ri}$, $I$ is an
$N_r \times N_r$ identity matrix, and therefore the size of $\Sigma$
is $MN_r \times MN_r$. The predictive distribution for the
interested variable $\textbf{f}^*$ on a new input $\textbf{x}^*$
which belongs to the $r$th component is
\begin{equation}
\mathcal{N} ( K^* \Sigma^{-1} \textbf{y}_r, \; K^{**} - K^*
\Sigma^{-1} K^{*\top}) ,
\end{equation}
where $K^*_{(M\times M N_r)}=\sigma_{r0}K_r \otimes
\textbf{k}_{r}^{x*}$, $K^{**} =\sigma_{r0}K_r$, and
$\textbf{k}_{r}^{x*}$ is a $1\times N_r$ row vector with the $i$th
element being $k_r(\textbf{x}^*,\textbf{x}_r^i)$. Hence, the
expected output on $\textbf{x}^*$ is $K^* \Sigma^{-1} \textbf{y}_r$.

Note that the calculation of $\Sigma^{-1}$ is a source for
approximation to speed up training. However, this problem is easier
than the original single GP model since we already reduced the
inversion from an $MN\times MN$ matrix to several $MN_r\times MN_r$
matrices.

We use hybrid Monte Carlo~\cite{Neal93MCMC} to update $\sigma_{r0}$,
and the basic Metropolis-Hastings algorithm to update $K_r$,
$\{w_{rd}\}$, and $\{\sigma_{r\ell}\}$ with the corresponding
proposal distributions being their priors. Below we give the
posteriors of the output space parameters and when necessary provide
some useful technical details.

We have
\begin{eqnarray}
&&p(\sigma_{r0}| Z,
\Theta\backslash \sigma_{r0},  \mathcal{D}) \nonumber\\
&\propto& p(\sigma_{r0})
p(\textbf{y}_r|\{\textbf{x}_{r}^{j}\}_{j=1}^{N_r}, \sigma_{r0}, K_r,
\{w_{rd}\}, \{\sigma_{r\ell}\}) \nonumber\\
&\propto& \sigma_{r0}^{a_1-1}\exp(-b_1 \sigma_{r0})\times
\frac{1}{|\Sigma|^{1/2}} \exp(-\frac{1}{2}\textbf{y}_r^\top
\Sigma^{-1} \textbf{y}_r)\nonumber\\
&=&\exp\Big\{-\Big[(1-a_1)\ln\sigma_{r0}+ b_1\sigma_{r0}+
\nonumber\\
&&\quad \frac{1}{2}\ln|\Sigma|+ \frac{1}{2} \textbf{y}_r^\top
\Sigma^{-1} \textbf{y}_r\Big]\Big\} ,
\end{eqnarray}
and thus the potential energy $E(\sigma_{r0})=(1-a_1)\ln\sigma_{r0}+
b_1\sigma_{r0}+\frac{1}{2}\ln|\Sigma|+ \frac{1}{2} \textbf{y}_r^\top
\Sigma^{-1} \textbf{y}_r$. The gradient $d E(\sigma_{r0})/ d
\sigma_{r0}$ is needed in order to use hybrid Monte Carlo, which is
given by
\begin{eqnarray}
&&\frac{d E(\sigma_{r0})}{d \sigma_{r0}}\nonumber\\
&=&\frac{1-a_1}{\sigma_{r0}} + b_1 +\frac{1}{2}
\mbox{Tr}[(\Sigma^{-1} - \Sigma^{-1}\textbf{y}_r
\textbf{y}_r^\top\Sigma^{-1})(K_r\otimes K_r^x)]. \nonumber
\end{eqnarray}

\begin{eqnarray}
\label{eqnCPKr} &&p(K_{r}| Z,
\Theta\backslash K_{r},  \mathcal{D}) \nonumber\\
&\propto& p(K_{r})
p(\textbf{y}_r|\{\textbf{x}_{r}^{j}\}_{j=1}^{N_r}, \sigma_{r0}, K_r,
\{w_{rd}\}, \{\sigma_{r\ell}\}) \nonumber\\
&\propto& |K_r|^{(\nu_1-M-1)/2}
\exp\{-\frac{1}{2}\mbox{Tr}(\mathbf{W}_1^{-1} K_r)\}
\times\nonumber\\
&& \frac{1}{|\Sigma|^{1/2}} \exp(-\frac{1}{2}\textbf{y}_r^\top
\Sigma^{-1} \textbf{y}_r)\nonumber\\
&=&\exp\Big\{-\frac{1}{2}\Big[(M+1-\nu_1)\ln|K_{r}|+
\mbox{Tr}(\mathbf{W}_1^{-1} K_r)+\nonumber\\
&&\quad\ln|\Sigma|+ \textbf{y}_r^\top \Sigma^{-1}
\textbf{y}_r\Big]\Big\} .
\end{eqnarray}

\begin{eqnarray}
&&p(w_{rd}| Z,
\Theta\backslash w_{rd},  \mathcal{D}) \nonumber\\
&\propto& p(w_{rd})
p(\textbf{y}_r|\{\textbf{x}_{r}^{j}\}_{j=1}^{N_r}, \sigma_{r0}, K_r,
\{w_{rd}\}, \{\sigma_{r\ell}\}) \nonumber\\
&\propto&  w_{rd}^{-1} \exp\{-\frac{(\ln w_{rd}-\mu_1)^2}{2
r_1}\}\times \frac{1}{|\Sigma|^{1/2}}
\exp(-\frac{1}{2}\textbf{y}_r^\top
\Sigma^{-1} \textbf{y}_r)\nonumber\\
&=&\exp\Big\{-\Big[\ln w_{rd}+ \frac{(\ln w_{rd}-\mu_1)^2}{2 r_1}
+ \nonumber\\
&&\quad \frac{1}{2}\ln|\Sigma|+\frac{1}{2} \textbf{y}_r^\top
\Sigma^{-1} \textbf{y}_r\Big]\Big\} .
\end{eqnarray}

\begin{eqnarray}
&&p(\sigma_{r\ell}| Z,
\Theta\backslash \sigma_{r\ell},  \mathcal{D}) \nonumber\\
&\propto& p(\sigma_{r\ell})
p(\textbf{y}_r|\{\textbf{x}_{r}^{j}\}_{j=1}^{N_r}, \sigma_{r0}, K_r,
\{w_{rd}\}, \{\sigma_{r\ell}\}) \nonumber\\
&\propto& \sigma_{r\ell}^{a_2-1}\exp(-b_2 \sigma_{r\ell})\times
\frac{1}{|\Sigma|^{1/2}} \exp(-\frac{1}{2}\textbf{y}_r^\top
\Sigma^{-1} \textbf{y}_r)\nonumber\\
&=&\exp\Big\{-\Big[(1-a_2)\ln\sigma_{r\ell}+
b_2\sigma_{r\ell}+\nonumber\\
&&\quad \frac{1}{2}\ln|\Sigma|+ \frac{1}{2} \textbf{y}_r^\top
\Sigma^{-1} \textbf{y}_r\Big]\Big\} .
\end{eqnarray}

\SubSection{Updating the concentration parameter $\alpha$}

The basic Metropolis-Hastings algorithm is used to update $\alpha$.
Let $1 \leq c \leq N$ be the number of distinct values in $\{z_1,
\ldots, z_N\}$. It is clear from~\cite{Antoniak74MDP} that
\begin{equation}
\label{eqnLikec} p(c|\alpha, N)=\beta^N_c \frac{\alpha^c \,
\Gamma(\alpha)}{\Gamma(N+\alpha)} ,
\end{equation}
where coefficient $\beta^N_c$ is the absolute value of Stirling
numbers of the first kind, and $\Gamma(\cdot)$ is the gamma
function. With  (\ref{eqnLikec}) as the likelihood, we can get the
posterior of $\alpha$ is
\begin{equation}
p(\alpha|c, N) \propto p(\alpha) p(c|\alpha, N) \propto p(\alpha)
  \frac{\alpha^c \, \Gamma(\alpha)}{\Gamma(N+\alpha)} .
\end{equation}
Since the gamma prior is used, it follows that,
\begin{equation}
\label{eqnPosAlpha} p(\alpha|c, N) \propto \frac{\alpha^{c+a_0-1}
\exp(-b_0\alpha) \Gamma(\alpha)}{\Gamma(N+\alpha)}.
\end{equation}

\Section{Prediction}
\label{sec:Prediction}

\begin{figure}[th]
  \centering
  \includegraphics[width=228pt]{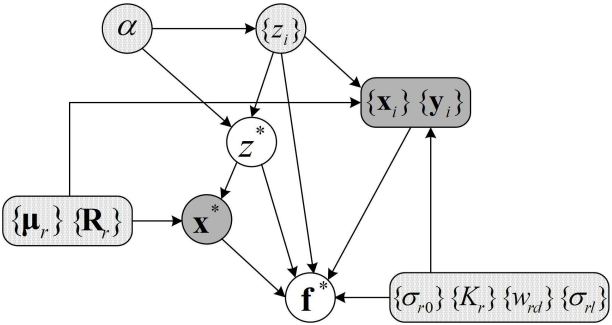}
  \caption{The graphical model for prediction on a new input $\textbf{x}^*$}
  \label{GM_IMMGP_Pred}
\end{figure}

The graphical model for prediction is shown in
Figure~\ref{GM_IMMGP_Pred}.

The predictive distribution for the predicted output of a new test
input $\textbf{x}^*$ is
\begin{eqnarray}
&&p(\textbf{f}^*|\textbf{x}^*, \mathcal{D}) =
\int \sum_Z \sum_{z^*} p(\textbf{f}^*, z^*, Z, \Theta|\textbf{x}^*, \mathcal{D}) d\Theta \nonumber\\
&=&\int \sum_Z \sum_{z^*} p(z^*, Z, \Theta|\textbf{x}^*,
\mathcal{D}) p(\textbf{f}^*| z^*, Z, \Theta, \textbf{x}^*, \mathcal{D}) d\Theta \nonumber\\
&=& \int \sum_{Z,z^*} p(z^*| Z, \Theta, \textbf{x}^*) p(Z,
\Theta|\textbf{x}^*,
\mathcal{D}) p(\textbf{f}^*| z^*, Z, \Theta, \textbf{x}^*, \mathcal{D}) d\Theta \nonumber\\
&\approx&  \int \sum_{Z, z^*} p(z^*| \textbf{x}^*, Z, \Theta) p(Z,
\Theta|\mathcal{D}) p(\textbf{f}^*| z^*, Z, \Theta, \textbf{x}^*,
\mathcal{D}) d\Theta \nonumber\\
&=&  \int \sum_Z \bigg[ \sum_{z^*} p(z^*| \textbf{x}^*, Z, \Theta)
p(\textbf{f}^*| z^*, Z, \Theta, \textbf{x}^*, \mathcal{D})
\bigg]\times\nonumber\\
&&\quad p(Z, \Theta|\mathcal{D}) d\Theta ,
\end{eqnarray}
where we have made use of the conditional independence $p(z^*| Z,
\Theta, \textbf{x}^*, \mathcal{D})=p(z^*| \textbf{x}^*, Z, \Theta)$
and a reasonable approximation $p(Z, \Theta|\textbf{x}^*,
\mathcal{D})\approx p(Z, \Theta|\mathcal{D})$.

With the Markov chain Monte Carlo samples $\{Z_i,
\Theta_i\}_{i=1}^L$ to approximate the above summation and
integration over $Z$ and $\Theta$, it follows that,
\begin{eqnarray}
&& p(\textbf{f}^*|\textbf{x}^*, \mathcal{D}) \nonumber\\
&\approx&\frac{1}{L} \sum_{i=1}^L \left[ \sum_{z^*} p(z^*|
\textbf{x}^*, Z_i, \Theta_i) p(\textbf{f}^*| z^*, Z_i, \Theta_i,
\textbf{x}^*, \mathcal{D}) \right] .\nonumber
\end{eqnarray}
Therefore, the prediction for $\textbf{f}^*$ is
\begin{equation}
\hat{\textbf{f}^*} = \frac{1}{L} \sum_{i=1}^L \left[ \sum_{z^*}
p(z^*| \textbf{x}^*, Z_i, \Theta_i) \mathbb{E}(\textbf{f}^*| z^*,
Z_i, \Theta_i, \textbf{x}^*, \mathcal{D}) \right],\nonumber
\end{equation}
where the expectation involved is simple to calculate since
$p(\textbf{f}^*| z^*, Z_i, \Theta_i, \textbf{x}^*, \mathcal{D})$ is
a Gaussian distribution, and $z^*$ takes values from $Z_i$ or is
different from $Z_i$ with the corresponding parameters sampled from
the priors.

The computation of $p(z^*| \textbf{x}^*, Z_i, \Theta_i)$ is given as
follows.
\begin{eqnarray}
&&  p(z^*| \textbf{x}^*, Z_i, \Theta_i) \nonumber\\
&=& \frac{p(z^*| Z_i, \Theta_i) p(\textbf{x}^*| z^*, Z_i,
\Theta_i)}{p(\textbf{x}^*|Z_i, \Theta_i)} \nonumber\\
&=& \frac{p(z^*| Z_i, \Theta_i) p(\textbf{x}^*| z^*, Z_i,
\Theta_i)}{\sum_{z^*} p(z^*| Z_i, \Theta_i) p(\textbf{x}^*| z^*,
Z_i, \Theta_i)} \nonumber\\
&=& \frac{p(z^*| Z_i, \Theta_i) p(\textbf{x}^*| z^*,
\Theta_i)}{\sum_{z^*} p(z^*| Z_i, \Theta_i) p(\textbf{x}^*| z^*,
\Theta_i)} ,
\end{eqnarray}
where the last equality follows from the conditional independence.
If $z^*=r \in Z_i$, then $p(z^*=r| Z_i,
\Theta_i)=\frac{N_{ir}}{\alpha + N}$ with $N_{ir}=\# \{z: z\in Z_i,
z=r\}$ and $p(\textbf{x}^*| z^*, \Theta_i)=p(\textbf{x}^*|\bm\mu_r,
\textbf{R}_r)$. If $z^* \notin Z_i$, then $p(z^*| Z_i,
\Theta_i)=\frac{\alpha}{\alpha + N}$ and
\begin{eqnarray}
p(\textbf{x}^*| z^*, \Theta_i) &=& \int p(\textbf{x}^*|\bm\mu,
\textbf{R})p(\bm\mu|\bm\mu_0,
\textbf{R}_0)p(\textbf{R}|\textbf{W}_0, \nu_0)d \bm\mu d \textbf{R}
.\nonumber
\end{eqnarray}
Unfortunately, this integral is not analytically tractable. A Monte
Carlo estimate by sampling $\bm\mu$ and $\textbf{R}$ from the priors
can be used to reach an approximation.

Note that, if $z^* \notin Z_i$, then $\mathbb{E}(\textbf{f}^*| z^*,
Z_i, \Theta_i, \textbf{x}^*, \mathcal{D})=\textbf{0}$ as a result of
zero-mean Gaussian process priors. Otherwise,
$\mathbb{E}(\textbf{f}^*| z^*, Z_i, \Theta_i, \textbf{x}^*,
\mathcal{D})$ can be calculated using standard Gaussian process
regression formulations.

\Section{Experiment}
\label{sec:Expe}

To evaluate the proposed infinite mixture model and the used
inference and prediction methods, we perform multivariate regression
on a synthetic data set. The data set includes 500 examples that are
generated by ancestral sampling from the infinite mixture model. The
dimensions for the input and output spaces are both set to two. From
the whole data, 400 examples are randomly selected as training data
and the other 100 examples serve as test data.

\SubSection{Hyperparameter setting} The hyperparameters for
generating data are set as follows: $a_0=1$, $b_0=1$,
$\bm\mu_0=\textbf{0}$, $\mathbf{R}_0=\mathbf{I}/10$,
$\mathbf{W}_0=\mathbf{I}/(10D)$, $\nu_0=D$,  $a_1=1$, $b_1=1$,
$\mathbf{W}_1=\mathbf{I}/M$, $\nu_1=M$, $\mu_1=0$, $r_1=0.01$,
$a_2=0.1$, and $b_2=1$. The same hyperparameters are used for
inference except $\boldsymbol{\mu}_0$, $\textbf{R}_0$ and
$\textbf{W}_0$. $\boldsymbol{\mu}_0$ and $\textbf{R}_0$ are set to
the mean $\boldsymbol{\mu}_x$ and inverse covariance $\textbf{R}_x$
of the training data, respectively. $\textbf{W}_0$ is set to
$\textbf{R}_x/D$.

\SubSection{Prediction Performance}

By Markov chain Monte Carlo sampling, we obtain 4000 samples where
only the last 2000 samples are retained for prediction. For
comparison purpose, the MTLNN approach (multitask learning neural
networks without ensemble learning)~\cite{Sun09MEL} is adopted.

Table~\ref{table_rmse} reports the root mean squared error (RMSE) on
the test data for our IMMGP model and the MTLNN approach. IMMGP1
only considers the existing Gaussian process components reflected by
the samples, while IMMGP2 considers to choose a new component as
well. The results indicate that IMMGP outperforms MTLNN and the
difference between IMMGP1 and IMMGP2 is very small.

\vspace{-0.3cm}
\begin{table}[ht]
\renewcommand{\arraystretch}{1.5}
\centering \caption{Prediction errors of different methods}
\vspace{0.2cm} \label{table_rmse}
\begin{tabular}{c|c|c}
\hline
MTLNN   & IMMGP1 & IMMGP2 \\
\hline
2.0659  & 0.7963 & 0.7964 \\
\hline
\end{tabular}
\end{table}
\vspace{-0.3cm}

\Section{Conclusion}
\label{sec:Conc}

In this paper, we have presented a new model called infinite
mixtures of multivariate Gaussian processes and applied it to
multivariate regression with good performance returned. Interesting
future directions include applying this model to large-scale data,
adapting it to classification problems and devising fast
deterministic approximate inference techniques.

%


\begin{thebibliography}{99}
\addtolength{\itemsep}{-0.8em}
\bibitem{Rasmussen06gpml}
C. Rasmussen and C. Williams, Gaussian Processes for Machine
Learning, MIT Press, Cambridge, MA, 2006.

\bibitem{Sun11viIMGP}
S. Sun and X. Xu, ``Variational inference for infinite mixtures of Gaussian processes
with applications to traffic flow prediction", IEEE Transactions on
Intelligent Transportation Systems, Vol. 12, No. 2, pp. 466-475,
2011.

\bibitem{Ji13M2SVM}
Y. Ji and S. Sun, ``Multitask multiclass support vector machines:
Model and experiments", Pattern Recognition, Vol. 46, No. 3, pp.
914-924, 2013.

\bibitem{Bonilla08MtGP}
E. Bonilla, K. Chai, and C. Williams,  ``Multi-task Gaussian process
prediction", Advances in Neural Information Processing Systems, Vol.
20, pp. 153-160, 2008.

\bibitem{Yuan11CMoR}
C. Yuan, ``Conditional multi-output regression", Proceedings of the
Interantional Joint Conference on Neural Networks, pp. 189-196,
2011.

\bibitem{Alvarez11CMoGP}
M. Alvarez and N. Lawrence, ``Computationally efficient convolved
multiple output Gaussian processes", Journal of Machine Learning
Research, Vol. 12, pp. 1459-1500, 2011.

\bibitem{Rasmussen02Infm}
C. Rasmussen and Z. Ghahramani,  ``Infinite mixtures of Gaussian
process experts", Advances in Neural Information Processing Systems,
Vol. 14, pp. 881-888, 2002.

\bibitem{Tresp01MGP}
V. Tresp,  ``Mixtures of Gaussian processes", Advances in Neural
Information Processing Systems, Vol. 13, pp. 654-660, 2001.

\bibitem{Teh10DP}
Y. Teh, ``Dirichlet processes", in Encyclopedia of Machine Learning,
Springer-Verlag, Berlin, Germany, 2010.

\bibitem{Meeds06IM}
E. Meeds and S. Osindero,  ``An alternative infinite mixture of
Gaussian process experts", Advances in Neural Information Processing
Systems, Vol. 18, pp. 883-890, 2006.

\bibitem{Bishop06prml}
C. Bishop, Pattern Recognition and Machine Learning,
Springer-Verlag, New York, 2006.

\bibitem{Rasmussen00IGMM}
C. Rasmussen,  ``The infinite Gaussian mixture model", Advances in
Neural Information Processing Systems, Vol. 12, pp. 554-560, 2000.

\bibitem{Neal98Sampling}
R. Neal, ``Markov chain sampling methods for Dirichlet process
mixture models", Technical Report 9815, Department of Statistics,
University of Toronto, 1998.

\bibitem{Neal93MCMC}
R. Neal, ``Probabilistic inference using Markov chain Monte Carlo
methods", Technical Report CRG-TR-93-1, University of Toronto, 1993.

\bibitem{Antoniak74MDP}
C. Antoniak, ``Mixtures of Dirichlet processes with applications to
Bayesian nonparametric problems", Annals of Statistics, Vol. 2, No.
6, pp. 1152-1174, 1974.

\bibitem{Sun09MEL}
S. Sun, ``Traffic flow forecasting based on multitask ensemble
learning", Proceedings of the ACM SIGEVO World Summit on Genetic and
Evolutionary Computation, pp. 961-964, 2009.

%
\end{thebibliography}

\end{document}